\documentclass[letterpaper, 10 pt, conference]{ieeeconf}  

\IEEEoverridecommandlockouts                              

\usepackage{graphics} 
\usepackage{amsmath} 
\usepackage{multirow}
\usepackage{makecell}
\usepackage{graphicx}
\let\labelindent\relax
\usepackage{enumitem}

\usepackage{ifthen,color,amssymb}
\newboolean{showcomments}
\setboolean{showcomments}{true} 
\ifthenelse{\boolean{showcomments}}
  {
		\newcommand{\nbb}[2]{
		\fcolorbox{black}{yellow}{\bfseries\sffamily\scriptsize#1}
		{\sf$\blacktriangleright$\textcolor{blue}{\textit{#2}}$\blacktriangleleft$}
		}

  }
  {
		\newcommand{\nbb}[2]{}
  }

\newcommand{\comment}[1]{}

\title{\LARGE \bf
Social-IWSTCNN: A Social Interaction-Weighted Spatio-Temporal Convolutional Neural Network for Pedestrian Trajectory Prediction in Urban Traffic Scenarios
}

\author{Chi Zhang,$^{1}$
        Christian Berger,$^{1}$
        Marco Dozza$^{2}$
\thanks{$^{1}$ Department of Computer Science and Engineering, University of Gothenburg, Gothenburg, Sweden.
        {\tt\small chi.zhang@gu.se, christian.berger@gu.se}}%
\thanks{$^{2}$ Department of Maritime Sciences and Mechanics, Chalmers University of Technology, Gothenburg, Sweden.
        {\tt\small marco.dozza@chalmers.se}}%
}

\begin{document}

\maketitle
\thispagestyle{empty}
\pagestyle{empty}

\begin{abstract}
Pedestrian trajectory prediction in urban scenarios is essential for automated driving. This task is challenging because the behavior of pedestrians is influenced by both their own history paths and the interactions with others. Previous research modeled these interactions with pooling mechanisms or aggregating with hand-crafted attention weights. In this paper, we present the Social Interaction-Weighted Spatio-Temporal Convolutional Neural Network (Social-IWSTCNN), which includes both the spatial and the temporal features. We propose a novel design, namely the Social Interaction Extractor, to learn the spatial and social interaction features of pedestrians. Most previous works used ETH and UCY datasets which include five scenes but do not cover urban traffic scenarios extensively for training and evaluation. In this paper, we use the recently released large-scale Waymo Open Dataset in urban traffic scenarios, which includes 374 urban training scenes and 76 urban testing scenes to analyze the performance of our proposed algorithm in comparison to the state-of-the-art (SOTA) models.
The results show that our algorithm outperforms SOTA algorithms such as Social-LSTM, Social-GAN, and Social-STGCNN on both Average Displacement Error (ADE) and Final Displacement Error (FDE). Furthermore, our Social-IWSTCNN is 54.8 times faster in data pre-processing speed, and 4.7 times faster in total test speed than the current best SOTA algorithm Social-STGCNN.
\end{abstract}

\section{INTRODUCTION}

Pedestrians are essential participants in urban traffic scenarios because they are vulnerable and need protection. Accurately predicting pedestrian trajectories is crucial for automated driving to reduce the risk of potentially hazardous situations. Recently, all major vehicle manufacturers strive to include more features towards increased levels of automated driving into cars. While many advanced driver-assistance systems (ADAS) on the Society of Automotive Engineers (SAE) Levels 1 and 2 are already commercially available, higher SAE levels, especially Level 4, require a more accurate prediction of the pedestrians than what is available today. 

The prediction of pedestrians' intentions is a great challenge because of the uncertainty and randomness of their motion pattern. Besides, as reported by Moussaid et al.~\cite{moussaid2010walking}, the moving behavior of pedestrians is not only dependent on their own paths from the past, but also driven by social interactions with others nearby.

Current state-of-the-art (SOTA) approaches aim at improving the accuracy of pedestrian trajectory prediction by using deep learning networks. Most related works learn the social interaction information by using a pooling module or a feature learning architecture over the hidden states learned from recurrent networks. However, Nikhil and Morris~\cite{nikhil2018convolutional} point out that trajectories are continuous in nature and do not have a complicated ``state''. Extracting social interactions from hidden states in previous models is indirect and the physical meaning of such hidden states is difficult to interpret. Instead of extracting features from hidden states, Social-STGCNN as reported by Mohamed et al.~\cite{mohamed2020social}, uses graph convolutional networks (GCNs) with a hand-crafted kernel to aggregate the features of other pedestrians directly from their position information to improve the prediction results.

While substantial improvements for predicting pedestrian trajectories were introduced by Social-STGCNN, its dependency on hand-crafted kernels with non-linear computations are time-consuming. Furthermore, the fixed kernel for social interaction aggregation could fail in modeling interactions between pedestrians correctly. Our research goal is to address these two challenges by presenting a novel structure to better capture the interactions between the pedestrians at a lower computational cost. We are looking in particular at the following two research questions:

\begin{description}
\item[RQ-1] To what extent can the proposed Social Interaction Extractor architecture learn the interaction weights in contrast to hand-crafted GCN kernels, to preserve and improve on the SOTA accuracy of predicting pedestrian trajectories while lowering the computational cost?
\item[RQ-2] How does our proposed network perform compared to the relevant SOTA approaches on the Waymo Open Dataset~\cite{sun2020scalability}, which provides 76 test scenes in densely populated urban traffic scenarios? 
\end{description}

In order to reach our research goal, we propose the Social Interaction-Weighted Spatio-Temporal Convolutional Neural Network (Social-IWSTCNN) to predict pedestrian trajectories in urban traffic scenarios. 

The main contributions of this paper are as follows:
\begin{itemize}
\item We propose a novel structure, the Social Interaction Extractor, 
to better and faster capture the interactions between pedestrians. Unlike the SOTA algorithm Social-STGCNN~\cite{mohamed2020social}, we avoid using the fixed attention weights with time-consuming non-linear computations in our approach, but learn the interaction attention weights with a data-driven approach. Compared with Social-STGCNN, our proposed network is 4.7 times faster in total test speed, and 54.8 times faster in data pre-processing speed, while producing competitive results.

\item As we aim to solve the real-world task of predicting the pedestrian trajectories in urban traffic scenarios, we use the recently released real-world large-scale Waymo Open Dataset for a systematic and experimental evaluation of our approach as this dataset contains more urban traffic scenarios that are nearly twice as densely populated by pedestrians than the datasets ETH~\cite{pellegrini2009you} and UCY~\cite{lerner2007crowds}, which were mainly used for performance comparisons so far. We compare our algorithm with three SOTA methods including Social-LSTM ~\cite{alahi2016social}, Social-GAN~\cite{gupta2018social}, and Social-STGCNN~\cite{mohamed2020social}, as well as other baseline methods like Linear Regression (LR) and long-short term memory (LSTM) models.
\end{itemize}
 
The remainder of the paper is structured as follows: Sec.~\ref{sec:RelatedWork} presents and discusses related work. Sec.~\ref{sec:Methodology} describes the proposed Social-IWSTCNN model. Sec.~\ref{sec:Experiments} presents the details of the experiments, and the results and findings thereof are provided in Sec.~\ref{sec:ResultsAnalysis}. Conclusions and future works are described in Sec.~\ref{sec:Conclusion}.

\section{RELATED WORK}
\label{sec:RelatedWork}
In recent years, the interest in intelligent vehicles and automated driving has attracted increasing attention to pedestrian trajectory prediction. Researchers in this field aim to deal with the interactions between pedestrians, as well as the predictions under uncertainty. Predicting pedestrians' intentions by hand-crafted features is difficult because their interactions are usually complicated and implicit. Recent research on deep learning has demonstrated the potential of learning the movement patterns of pedestrians in a data-driven manner. In this section, we provide a brief summary and discussion of related work.
\subsection{RNN-based Pedestrian Trajectory Prediction}
LSTM, an improved version of recurrent neural networks (RNNs), is preferred by many researchers because of its strong ability to handle the trajectory sequence information. Social-LSTM as proposed by Alahi et al.~\cite{alahi2016social} uses LSTMs to extract the movement feature state from each pedestrian, and applies a ``social pooling'' layer over LSTMs to model social interactions. The Social-LSTM model assumes that the pedestrian trajectories follow bi-variate Gaussian distributions and predicts probabilistic trajectories. In this paper, we follow the bi-variate Gaussian distribution assumption and use negative log likelihood as the loss function. 

Later works have extended Social-LSTM methods by improving the interaction aggregation module or by including environment and neighbor features extracted from images. Fernando et al.~\cite{fernando2018soft+} apply a ``soft'' and ``hard-wired'' combined attention model over LSTMs to improve the prediction precision. State-Refinement LSTM~\cite{zhang2019sr} improves the pooling mechanism to better capture the interactions between pedestrians by refining the states of pedestrians. Xue et al.~\cite{xue2018ss} further include scene information to the LSTM-based framework. Their proposed Social-Scene-LSTM uses two additional LSTMs to tackle neighboring pedestrians and scene layout information and predicts deterministic trajectories by minimizing the least square error. Chandra et al.~\cite{chandra2019traphic} use convolutional neural networks (CNNs) to extract the features from images separately from neighbor and horizon views, and concatenate them together with ego-state to be fed into the LSTM framework. These approaches have gained good performance on pedestrian trajectory prediction.

Some researchers argue that multiple trajectories are plausible and socially-acceptable given a trajectory from the past, and assume that the pedestrian trajectories follow a multi-modal distribution. For instance, Gupta et al.~\cite{gupta2018social} propose Social-GAN using generative adversarial networks (GANs) with an LSTM based generator, and other researchers follow this direction. Social-ways as reported by Amirian et al.~\cite{amirian2019social} improves the attention pooling structure instead of max pooling and uses an info-GAN without L2 loss. SoPhie as proposed by Sadeghian et al.~\cite{sadeghian2019sophie} improves the attention mechanism and includes the environment scene feature extracted from images by CNNs within the LSTM-based GAN framework. CGNS~\cite{li2019conditional} is similar to SoPhie but uses gate recurrent units (GRUs) instead of LSTMs for prediction to improve the accuracy. The approach Social-BiGAT as described by Kosaraju et al.~\cite{kosaraju2019social} is also based on GANs but improves the social attention learning module with a graph attention network over the hidden states of LSTMs to model social interactions.

\subsection{CNN-based Pedestrian Trajectory Prediction}
In addition to extracting features from images, CNNs have also been used for trajectory prediction. Unlike RNN methods whose later time-step prediction depends on previously predicted time-steps, CNN-based methods can predict all time-steps at once. Nikhil and Morris~\cite{nikhil2018convolutional} use CNNs to predict the trajectories and achieve competitive results while reaching computational efficiency. Bai et al.~\cite{bai2018empirical} argue that the parameters of RNNs are inefficient and expensive in training, and hence, the usage of Temporal 
Convolutional Networks (TCNs) should be considered for sequence modeling tasks. Mohamed et al.~report Social-STGCNN~\cite{mohamed2020social}, which has made a great breakthrough on reducing the final error by using TCNs and CNNs yet with a faster speed. Therefore, in this paper we use TCNs and CNNs for long-term trajectory predictions.

\subsection{Social Interaction Extracting}
How to learn the social interactions is one of the most concerned topics in pedestrian trajectory prediction. Pioneering work from Helbing and Molnar~\cite{helbing1995social} presents the Social Force model to handle interactions with attractive and repulsive forces, but the hand-crafted rules are hard to generalize. Alahi et al.~\cite{alahi2016social} propose a ``social pooling'' layer, which uses a rectangular shape occupancy grid map of the neighborhood to represent the relationship of neighbors. Gupta et al.~\cite{gupta2018social} point out that local information is not always sufficient, and hence, they use a multi-layer perceptron (MLP) followed by a max pooling structure to capture the global social interaction information. Sadeghian et al.~\cite{sadeghian2019sophie} assume that people pay more attention to closer objects so they sort the attention by distance.
GCNs are defined as the convolution operation over graphs~\cite{kipf2016semi}, which are weighted aggregations of target nodes with the features of their neighbor nodes.
As the above mentioned hand-crafted modules may fail in learning the interactions correctly, Social-BiGAT~\cite{kosaraju2019social} uses GCNs as graph attention networks to extract the social interactions between pedestrians. 

Most of the previous models have applied the social interaction layer on hidden states extracted from recurrent networks, expecting such hidden states to capture the pedestrians' motion properties. However, trajectories are continuous in nature and do not have complicated ``states''~\cite{nikhil2018convolutional}. The physical meaning of hidden states is difficult to interpret, and using these ``states'' to represent the motion properties is indirect and non-intuitive. Mohamed et al.~\cite{mohamed2020social} propose the Social-STGCNN that learns the embedding feature of pedestrians directly from their locations, and apply GCNs with a hand-crafted kernel as the weights to aggregate the feature from other pedestrians.

However, the hand-crafted kernel of Social-STGCNN includes two square computations and one square root computation for each pair of pedestrians, and these non-linear computations are very time-consuming. Furthermore, although the fixed kernel has some physical meaning and can be explained, it does not learn the interaction information from the data, and may represent the social interaction relationships incorrectly. Therefore, in this paper we do not spend long time on building graph. Instead, we propose a novel design to \emph{learn} the social interaction weights between pedestrians, which avoids the time-consuming non-linear computations but learns such social interaction attention weights from their relative positions in a data-driven manner.

\subsection{Datasets}
\label{sec: dataset}
High-quality and large-scale datasets are crucial for data-driven machine learning algorithms.
Two publicly available datasets ETH~\cite{pellegrini2009you} and UCY~\cite{lerner2007crowds} contain bird's-eye-view videos and image annotations of pedestrians collected in real-world scenarios. Most of previous pedestrian trajectory prediction algorithms are trained and evaluated on these datasets ~\cite{alahi2016social,zhang2019sr,xue2018ss,gupta2018social,sadeghian2019sophie,li2019conditional,kosaraju2019social,mohamed2020social}. There are in total five scenes in these datasets, containing 6,441 frames at 2.5~Hz. The maximum number of pedestrians in each frame is 75 and the average number of pedestrians is 14.6.

However, both ETH and UCY datasets do not include sufficiently enough urban traffic scenarios for a detailed analysis of relevant, densely populated urban traffic situations. Recently, Waymo released a real-world large-scale dataset ~\cite{sun2020scalability}, which is the largest and most diverse autonomous vehicle dataset ever published, including urban and suburban scenarios in high quality. This dataset consists of 1,150 scenes in total, 450 scenes of which are urban scenarios. Each of the scene has a duration of 20 seconds, containing LiDAR and camera data with multiple objects labeled by 2-dimensional (2D) and 3-dimensional (3D) bounding boxes and unique track identifiers. The urban scenarios in the dataset consist of 374 training scenes and 76 test scenes. The dataset contains in total 20,697 frames at 2.5~Hz. The maximum number of pedestrians in each frame is 195 and the average number of pedestrians is 27.4. Since the Waymo Open Dataset is 3.21 times larger and 1.85 more densely populated than the ETH and UCY datasets, it is more appropriate to evaluate the models on Waymo Open Dataset. In this paper, we train and evaluate our algorithm comparing to the SOTA models using Waymo Open Dataset.

\section{Methodology}
\label{sec:Methodology}
\subsection{Problem Definition}
\label{ProblemDefination}
A person's position in a scene is represented by real-world x-y-coordinate $X=(x, y)$. Given a set of $n$ pedestrians with their observed positions $X_t^i=(x_t^i, y_t^i)$ where $i\in{\{1,\dots,n\}}$, over time-steps $1 \leq t \leq T_{obs}$, we aim to predict the likely trajectories of pedestrians $\hat Y_t^i=(\hat x_t^i, \hat y_t^i)$ in the future time-steps $T_{obs}+1 \leq t \leq T_{pred}$. The ground truth of the future trajectories is denoted as $Y_t^i=(x_t^i, y_t^i)$, where $i\in{\{1,...,n\}}, T_{obs}+1 \leq t \leq T_{pred}$.

The predicted positions of pedestrians $\hat Y_t^i=(\hat x_t^i, \hat y_t^i)$, $i\in{\{1,...,n\}}, T_{obs}+1 \leq t \leq T_{pred}$ are random variables. We assume that the $i^{th}$ pedestrian's position at time $t$ follows bi-variate Gaussian distribution $\hat Y_t^i \sim \mathcal{N}(\mu_t^i,\,\sigma_t^i,\,\rho_t^i)$. At time-step $t$, the mean value of the position is $\mu_{t}^i = (\mu_x, \mu_y)_{t}^i$. The standard deviation is $\sigma_{t}^i = (\sigma_x, \sigma_y)_{t}^i$, and the correlation coefficient is $\rho_{t}^i$. To get the trajectory prediction, our network predicts the Gaussian distribution parameters $(\mu_x,\mu_y,\sigma_x,\sigma_y,\rho)_t^i$. 

The pedestrian positions of observation time-steps $1 \leq t \leq T_{obs}$ are used for predicting the distribution of the pedestrian trajectory positions during prediction time-steps $T_{obs}+1 \leq t \leq T_{pred}$.
We use the negative log likelihood loss function to learn the parameters of the model, as below:
\begin{equation}
\label{eq_loss}
L(W) = -\sum_{t=T_{obs}+1}^{T_{pred}}\sum_{i=1}^{n}\log{(f({x_t^i,y_t^i|\mu_x,\mu_y,\sigma_x,\sigma_y,\rho}))}
\end{equation}
where $W$ represents the learned network parameters. We minimize the loss value to get optimal weights for our network.

\subsection{Overall Model}
In this part, we present the Social-IWSTCNN algorithm that learns the social interaction weights between pedestrians. We designed this approach for dense traffic scenarios in the city and used the Waymo Open Dataset for training and evaluation. Our proposed network takes pedestrians' sequences with x-y-coordinate center positions from bird's-eye-view as inputs, and generates the predicted trajectories of pedestrians.

\begin{figure*}
\begin{center}
\includegraphics[scale=0.75]{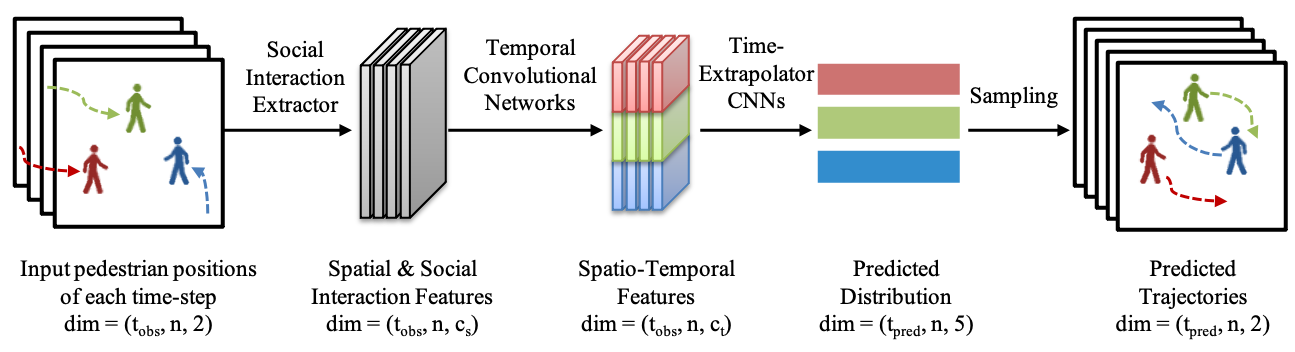}
\end{center}
  \caption{Overall framework of Social-IWSTCNN. Given observed frame sequences, we use the positions in each frame as input to learn the social interaction weights, and extract spatial and social interaction features using Social Interaction Extractor. Following this, we apply TCNs to create spatio-temporal features for each pedestrian. Then we apply Time-Exgrapolator CNNs to predict future trajectory distributions. Finally we sample to get the predicted trajectories.}
\label{fig:framework}
\end{figure*}

The overall Social-IWSTCNN model mainly includes three parts: a)~the Social Interaction Extractor to learn the interaction weights and spatial features, b)~the Temporal Convolutional Networks for temporal feature extracting, and c)~the Time-Extrapolator Convolutional Network for prediction. The overall network structure is shown in Fig.~\ref{fig:framework}.

Both the spatial and the temporal features are included in the network. The spatial features are extracted by the Social Interaction Extractor, including the input position embedding feature as well as the social interaction feature. The temporal features are learned by TCNs, which use convolution operations over the sequential spatial features to capture the temporal patterns of the sequence. Finally, after extracting the spatial and temporal features of each pedestrian, a CNN is used for extrapolating the long-term trajectories. The details are shown in the following sections. 

\subsection{Feature Embedding and Social Interaction Extractor}
\begin{figure}
\begin{center}
\includegraphics[scale=0.52]{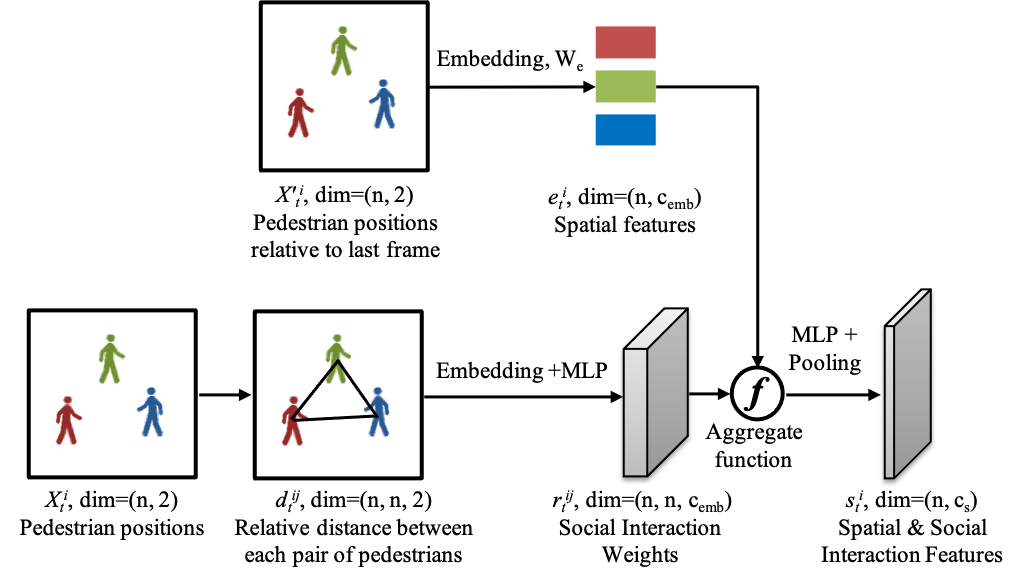}
\end{center}
  \caption{Pedestrian Social Interaction Extractor. The input are the relative positions to last frame and pedestrian positions of each time-step. We use MLP to learn the social interaction weights, and use an aggregate function to extract the spatial and social interaction features.}
\label{fig:interaction}
\end{figure}

This module is used to extract the spatial features and to capture the social interactions between pedestrians. The process is shown in Fig.~\ref{fig:interaction}.

Because of the limitation of using hidden states of LSTMs to represent the human motion property, we propose a structure to learn the feature of pedestrians from their positions directly. 
As outlined earlier, we found that the graph representation in the Social-STGCNN model consumes a considerable amount of time to build the graphs. This is mainly caused by the edges' attribute calculation:
$a^{ij}_t=({(x^i_t-x^j_t)^2+(y^i_t-y^j_t)^2})^{-\frac{1}{2}}$
, which includes two square computations and one square root computation for each pair of pedestrians $i,j$ in a single time-step $t$, and these non-linear computations are very time-consuming. Furthermore, building the spatial graph $G_t=(V_t, E_t)$ needs to copy the already existing observed relative locations to vertices $V_t$, which is not necessary and resource-consuming.

Therefore, in our model, we do not build a graph representation of pedestrian trajectories. Instead, we directly use the observed locations relative to last frame at each time-step as input for feature capturing. The spatial features of pedestrian $i$ at time-step $t$ are captured by embedding the input x-y-coordinate positions as shown in:
\begin{equation}
e_t^i = {X_t^i}{W_{e}}
\label{eq_input_emb}
\end{equation}
where the embedded features are denoted as $e_t^i$. The embedding weights are $W_{e}$, and $X_t^i$ are input trajectories.

As pedestrians adjust their paths implicitly depending on their neighbors, we introduce the Social Interaction Extractor to learn the social interaction relationship between pedestrians after getting their embedded spatial features. The input of the Social Interaction Extractor is the relative distances between pedestrians. We use linear embedding followed by an MLP to learn the interactions, and the outputs are the interaction attention weights between pedestrians, as denoted in:
\begin{equation}
r_t^{ij} = MLP(d_t^{ij}W_{r}; W_{s}), i \neq j
\label{eq_social}
\end{equation}
where $r_t^{ij}$ stands for the attention weights between pedestrians $i$ and $j$ at time-step $t$. It is the indicator of how much the $i^{th}$ pedestrian will be affected by the $j^{th}$ pedestrian. The $MLP (\cdot)$ represents the MLP operation using PreLU~\cite{he2015delving} as the non-linear activation function. $d_t^{ij}$ are the relative positions between $i^{th}$ and $j^{th}$ pedestrian, $d_t^{ij} = (\Delta x^{ij}, \Delta y^{ij})=(x^j - x^i, y^j - y^i)$. $W_r$ are linear embedding weights, and $W_s$ are the learned parameters of the MLP.

Then we use the learned social interaction attention weights $r_t^{ij}$ to multiply the spatial features $e_t^i$ and aggregate the spatial social feature $f_t^i$ as shown below:
\begin{equation}
f_t^i=\sum_{j\in{n}} {e_t^j} \cdot {r_t^{ij}}, i \neq j
\label{eq_social_interaction1}
\end{equation}
Finally, an MLP and a pooling layer are applied to extract the spatial and social interaction features, and to get the state features into a tensor, and then pass to the TCN. The function is shown in
\begin{equation}
s_t^i=Pooling\{MLP(f_t^{i}; W_c)\}
\label{eq_social_interaction}
\end{equation}
where $s_t^i$ is the final embedded spatial social interaction feature of pedestrian $i$ at time $t$, $f_t^{i}$ is the spatial social feature, and $W_c$ are the learned parameters of the MLP.
\subsection{The TCN and CNN for Time Series Prediction}
Inspired by Social-STGCNN~\cite{mohamed2020social}, we use the TCN and CNN-based approach to process the pedestrian trajectories. By applying the convolutional operations on the temporal space, we can efficiently capture the temporal relationship in time-scales for sequential predictions and alleviate the error accumulating problem caused by RNNs. After extracting the temporal features, the CNN extrapolator is used to directly predict all time-steps of the prediction horizon at once.
\section{Experiments}
\label{sec:Experiments}
\subsection{Dataset Introduction and Pre-processing}
In this section, we train and evaluate our algorithm on Waymo Open Dataset~\cite{sun2020scalability}, because it contains more sufficient urban traffic scenarios than the previously used ETH~\cite{pellegrini2009you} and UCY~\cite{lerner2007crowds} datasets
as we discussed before in Sec.~\ref{sec: dataset}.

The original record sequences of the Waymo Open Dataset~\cite{sun2020scalability} have a frequency of 10~Hz. The algorithms we compare with are previously evaluated on ETH and UCY datasets and sampled at 2.5~Hz. To compare the results with the SOTA models, we keep the same frequency in this paper and sample the sequences of Waymo Open Dataset to 2.5~Hz. The pedestrians are labeled by 3D bounding boxes on LiDAR data with their real-world center position $(x, y, z)$ and size $(length, width, height)$. The scan range of the LiDAR sensor is 75 m, and we use all labeled pedestrians in scan range. Our algorithm takes 2D pedestrian position $(x, y)$ sequences as input. Therefore, we use a 2D bird's-eye-view image map representation with x-y-coordinate center positions, and stored them as sequences with unique track identifiers. The objects are taken as points without size information in this paper. We store the data in four columns: frame id, pedestrian track id, x position, and y position.

The Waymo Open Dataset uses local coordinates with ego-vehicle center as the origin. However, this will introduce the movement of ego-vehicle into the pedestrians' movement. To solve this problem, we processed the coordinates and transformed them to global coordinates. The origin of each record sequence is set to the ego-vehicle's position of the first time-step of the record. In the data loading module of the training and testing part, we cut the sequence into pieces with a fixed sequence length. We set the sequence length as the sum of the observation length and prediction length. We set the sequence length at 20 time-steps with 8 observation time-steps (ie., 3.2 seconds) and 12 prediction time-steps (ie., 4.8 seconds), and get 195,192 training sequences, 36,946 validation sequences, and 52,484 evaluation sequences, which is large and sufficient for training and evaluating.

To evaluate the algorithms in different crowd densities, we divided the test set into three groups by the average pedestrian number per frame. The information of the three groups are listed in Table \ref{table: data}. We looked into the scenarios of each group. In Group 1, the scenarios are mainly curve-roads and irregularly shaped roads containing very few people. In contrast, in Groups 2 and 3, the scenes are mainly crossroads and intersections with pedestrians, and Group 3 is more densely populated with crowds compared with Group 2.
\begin{table}[h]
    \begin{center}
    \caption{Test Dataset Description}
    \label{table: data}
    \begin{tabular}{c||c|c|c|c}
        \hline
        \makecell{Dataset \\ group} & Description & \makecell{Avg. pedestrian \\ number \\ per frame, n} & \makecell{Number \\ of scenes} & \makecell{Number of \\ sequences} \\
        \hline
        \hline
        Group 1 & Least crowded & $n<15$ & 30 & 2,005 \\
        \hline
        Group 2 & \makecell{Moderate \\ crowded} & $18<n<62$ & 33 & 24,489 \\
        \hline
        Group 3 & Most crowded & $n>71$& 13 & 25,990 \\
        \hline
        \hline
        In total & \makecell{The whole \\ dataset} & $0<n<160$ & 76 & 52,484 \\
        \hline
    \end{tabular}
    \end{center}
\end{table}

\subsection{Evaluation Metrics and Baselines}
\label{sec:EvaluationMetrics}
We use two metrics to report the prediction error: 
\begin{itemize}
\item The Average Displacement Error (ADE): the average distance between ground truth and prediction trajectories over all predicted time-steps, as defined below:
\begin{equation}
ADE=\frac{\sum_{i\in{n}}\sum_{t=T_{obs}+1}^{T_{pred}}{\| Y_t^i - \hat Y_t^i \|}_2}{n \times (T_{pred} - T_{obs})}
\label{eq_ade}
\end{equation}

\item The Final Displacement Error (FDE): the average distance between ground truth and prediction trajectories for the final predicted time-step, as defined below:
\begin{equation}
FDE=\frac{\sum_{i\in{n}}{\| Y_t^i - \hat Y_t^i \|}_2}{n }, t = T_{pred}
\label{eq_fde}
\end{equation}
\end{itemize}

In our algorithm, we only use the trajectory information, and hence, all models we selected for comparison do not use other information such as scene images, camera, or LiDAR information. We compare the performance of our proposed models against the following baseline methods:
\begin{enumerate}[label=(\alph*)]
\item Linear Regression (LR): A linear regression model of pedestrian motion over each dimension.
\item LSTM: Na\"ive LSTM without the influence of other individuals.
\item Social-LSTM: as proposed by Alahi et al.~\cite{alahi2016social} in 2016.
\item Social-GAN: as proposed by Gupta et al.~\cite{gupta2018social} in 2018.
\item Social-STGCNN: as proposed by Mohamed et al.~\cite{mohamed2020social} in 2020.
\item Social-IWSTCNN (ours): our proposed method as described in Sec.~\ref{sec:Methodology}. 
\end{enumerate}

\subsection{Implementation Details}
We trained Social-IWSTCNN model with the Stochastic Gradient Decent (SGD) with an initial learning rate 0.01. The training batch size is set to 64 for 250 epochs. We used Nvidia GeForce RTX 2080 Ti GPU for our training and evaluating. The hyper-parameters are determined empirically. Since the Social-IWSTCNN model predicts the bi-variate Gaussian distribution, we follow the evaluation method used in Social-STGCNN, which generates 20 samples and uses the closest sample to the ground truth for metrics computation. For Social-GAN model, we generate 20 samples and use the closest for evaluation. For both LR and LSTM models, we get single prediction and it is used for evaluation.

To better evaluate the previous methods, all comparing models are re-trained on the Waymo Open Dataset. We use the first 8 time-steps covering 3.2 seconds for observation to predict the last 12 time-steps covering 4.8 seconds.
\section{Results and Analysis}
\label{sec:ResultsAnalysis}
\subsection{Quantitative Evaluation}
The results of all methods mentioned in Sec.~\ref{sec:EvaluationMetrics}
are compared on the Waymo Open Dataset~\cite{sun2020scalability}, and the ADE and FDE are shown in Table~\ref{table:results1}. The results are for 12 time-steps (ie., 4.8 seconds) prediction. The ADE and FDE are in meters, the lower the better. Overall, Social-IWSTCNN outperforms all previous methods on the two metrics.

\begin{table*}[h]
    \begin{center}
    \caption{The ADE/FDE metrics for several methods compared to Social-IWSTCNN in different test groups.}
    \label{table:results1}
    \begin{tabular}{c|c|c|c|c|c|c||c}
        \hline
        Metric & Dataset & LR & LSTM & Social-LSTM & Social-GAN & Social-STGCNN & Social-IWSTCNN   \\
        (Year) & Group & & & (2016) & (2018) & (2020) & (Ours) \\
        \hline
        \hline
        & Group 1 & 0.500 & 0.495 & 0.474 & 0.422 & 0.509&\bf{0.421} \\ \cline{2-8}
        ADE & Group 2 & 0.400 & 0.397 & 0.403 & 0.393 & 0.338 & \bf{0.335}  \\ \cline{2-8}
        & Group 3 & 0.417 & 0.370 & 0.395 & 0.375 & 0.318 & \bf{0.314}  \\ \cline{1-8}
        \bf{Average} & The whole dataset & 0.412 & 0.384 & 0.402 & 0.386 & 0.334 & \bf{0.329} \\
        \hline
        \hline
        & Group 1 & 1.091 & 1.095 & 1.030 & 0.908 & 0.907 & \bf{0.755} \\ \cline{2-8}
        FDE & Group 2 & 0.879 & 0.852 & 0.844 & 0.838 & 0.561 & \bf{0.559} \\ \cline{2-8}
        & Group 3 & 0.888 & 0.793 & 0.818 & 0.806 & 0.511 &\bf{0.499} \\ \cline{1-8}
        \bf{Average} & The whole dataset & 0.892 & 0.829 & 0.840 & 0.826 & 0.550 &\bf{0.540} \\
        \hline
    \end{tabular}
    \end{center}
\end{table*}
 
The Social-IWSTCNN gets competitive results compared with the previous SOTA model Social-STGCNN on the whole test set. We notice that on Group 2 and Group 3 which are more densely populated, the results of Social-IWSTCNN are only slightly better than Social-STGCNN, while on Group 1 which is less densely populated, ADE is substantially reduced by 17.3\% and FDE is reduced by 16.8\%, resulting in a significant improvement using our Social-IWSTCNN. This shows that, in a less dense scenario, the proposed Social Interaction Extractor can better capture the social interactions between pedestrians than the hand-crafted interaction aggregation weights in Social-STGCNN. 

Both the Social-IWSTCNN and Social-STGCNN get more accurate results than the LSTM-based methods~\cite{alahi2016social,gupta2018social}. The reasons are as we outlined before: Firstly, the hidden states of LSTMs could not represent the movement property as good as embedding the feature directly from the spatial and temporal information; secondly, the recurrent prediction will accumulate the error and therefore, the FDE of LSTM-based methods become larger than the CNN-based methods. Furthermore, we notice that for the Social-LSTM and Social-GAN methods, ADE and FDE of Group 3 are worse than the LSTM results. This is an evidence that the pooling structure on the hidden states of LSTMs cannot extract the interaction feature properly in dense urban traffic scenarios. 

To our surprise, for all deep learning-based algorithms, the ADE and FDE on Group 1 perform worse than on Group 2 and 3. This shows that the less crowded scenarios do not mean that they are easier to predict. When there are only a few pedestrians, they may tend to interact with the environment scene and the vehicles. Without adding other information, the scenarios with less pedestrians are difficult to predict. By contrast, the results of LR method are less sensitive to crowd density.

{\bf Inference speed}: We compared the inference speed between the two competitive methods: Social-IWSTCNN and Social-STGCNN. The time consumption and speed up are listed in Table \ref{table: time}.
Social-STGCNN has an inference time of 3.20~ms per sequence, but the pre-processing including the graph building time takes 12.61~ms per sequence. The total inference time is 15.81~ms. By contrast, our proposed method takes 3.15~ms inference time per sequence, and 0.23~ms data pre-processing time per sequence, which is 54.8 times faster than Social-STGCNN, and the total inference time is 3.38~ms, which is 4.7 times faster.

In our algorithm, we remove the non-linear calculation for attention weights computing, and avoid constructing adjacent matrix of the graph. To evaluate the influence of the two changes, we test the inference time of the Social-IWSTCNN method with graph construction to see how much our method can speed up by only removing the non-linear calculation. The results show that by removing the non-linear calculation, the pre-processing time pre sequence is 2.90~ms, which is 4.3 times faster than Social-STGCNN, and total inference time is 5.83~ms, which is 2.7 times faster. This result shows that both removing of non-linear calculation and avoiding graph constructing improves the inference speed. Our algorithm is computationally more efficient and has a much better performance on speed while reaching competitive results.  
\begin{table}[h]
    \begin{center}
    \caption{Inference time comparison (speed up).}
    \label{table: time}
    \begin{tabular}{c||c|c|c}
        \hline
        Time & Pre-processing & Inference & Total \\
        \hline
        \hline
        Social-STGCNN & 12.61 & 3.20 & 15.81 \\
        \hline
        \makecell{Social-IWSTCNN (with \\ graph construction)} & 2.90 (x4.3) & 2.93 & 5.83 (x2.7) \\
        \hline
        Social-IWSTCNN & 0.23 \bf{(x54.8)} & 3.15 & 3.38 \bf{(x4.7)} \\
        \hline
    \end{tabular}
    \end{center}
\end{table}

{\bf Long-term prediction:} We predicted long-term trajectories and the results are shown in Table \ref{table: long-term}. Both Social-STGCNN and Social-IWSTCNN outperform Social-GAN when the prediction time expands to 8.0~seconds (ie., 20 time-steps). As we previously discussed, the LSTM-based methods will accumulate the error because the later steps base on the previous output results.
The results show that the CNN-based methods (Social-STGCNN and Social-IWSTCNN) are more suitable for long-term predictions.
\begin{table}[h]
    \begin{center}
    \caption{Long-term prediction, ADE/FDE.}
    \label{table: long-term}
    \begin{tabular}{c||c|c}
        \hline
        Model & 4.8 seconds (12 steps) & 8.0 seconds (20 steps) \\
        \hline
        \hline
        Social-GAN & 0.386 / 0.826 & 0.773 / 1.684 \\
        \hline
        Social-STGCNN & 0.334 / 0.550 & \bf{0.650} / \bf{1.212} \\
        \hline
        Social-IWSTCNN & \bf{0.329} / \bf{0.540} & 0.654 / 1.213 \\
        \hline
    \end{tabular}
    \end{center}
\end{table}

\subsection{Qualitative Evaluation}
\begin{figure*}
\begin{center}
\includegraphics[scale=0.43]{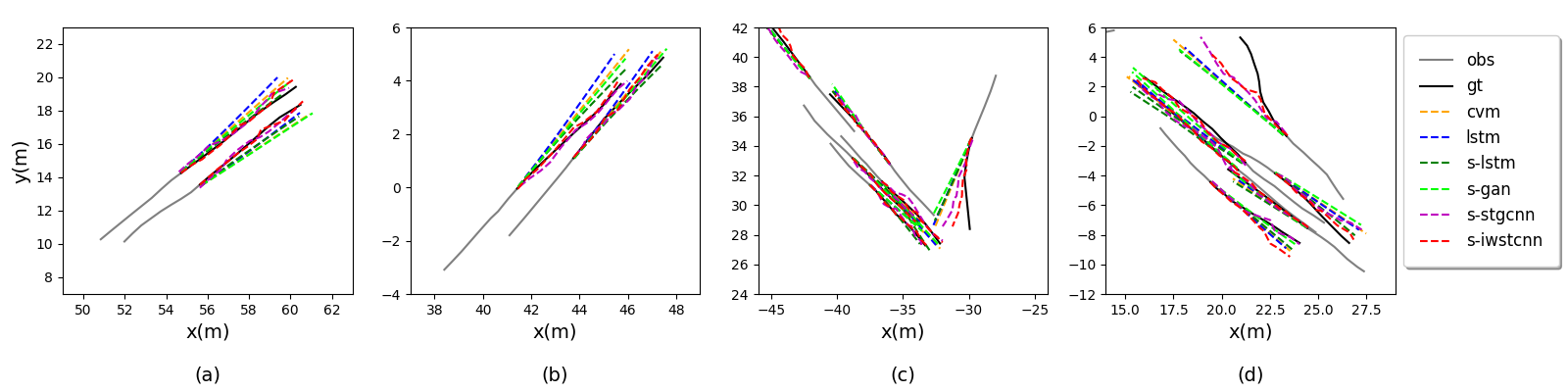}
\end{center}
  \caption{The comparison of prediction results of different algorithms in various scenarios. (a) Two individuals walking in parallel, (b) two individuals walking in parallel and merging, (c) collision avoidance, the individual from the top right corner tend to avoid the pedestrians on the bottom left corner, and (d) individuals from the top left corner meeting a group from the bottom right corner. The legends: obs stands for observed paths; gt stands for the ground truth of predicted trajectories. s-lstm stands for Social-lstm; s-gan stands for Social-GAN; s-stgcnn stands for Social-STGCNN; and s-iwstcnn stands for our proposed method Social-IWSTCNN.}
\label{fig:figure3}
\end{figure*}

Next, we qualitatively analyze the performance of Social-IWSTCNN and compare it with other methods. In Fig.~\ref{fig:figure3}, we show the results in different scenarios. Fig.~\ref{fig:figure3}(a) shows the scenario that two individuals are walking in parallel, and Fig.~\ref{fig:figure3}(b) is similar to Fig.~\ref{fig:figure3}(a) but the two individuals are merging. In these two scenarios, the Social-IWSTCNN and Social-STGCNN perform better than the other LSTM-based methods. The two parallel pedestrians tend to interact with each other and our proposed method managed to better capture their interactions. Fig.~\ref{fig:figure3}(c) shows a collision avoidance scenario: The pedestrians from the top right corner tend to avoid the pedestrians in bottom left corner. The Social-IWSTCNN predicted this trend, while the LSTM-based methods failed to avoid the collision. This shows that Social-IWSTCNN performs better on capturing the social interaction feature between pedestrians. Fig.~\ref{fig:figure3}(d) shows how pedestrians react when individuals meeting a group: In the prediction, Social-IWSTCNN successfully avoids the collision and the group of pedestrians keeps walking together.

\begin{figure*}
\begin{center}
\includegraphics[scale=0.43]{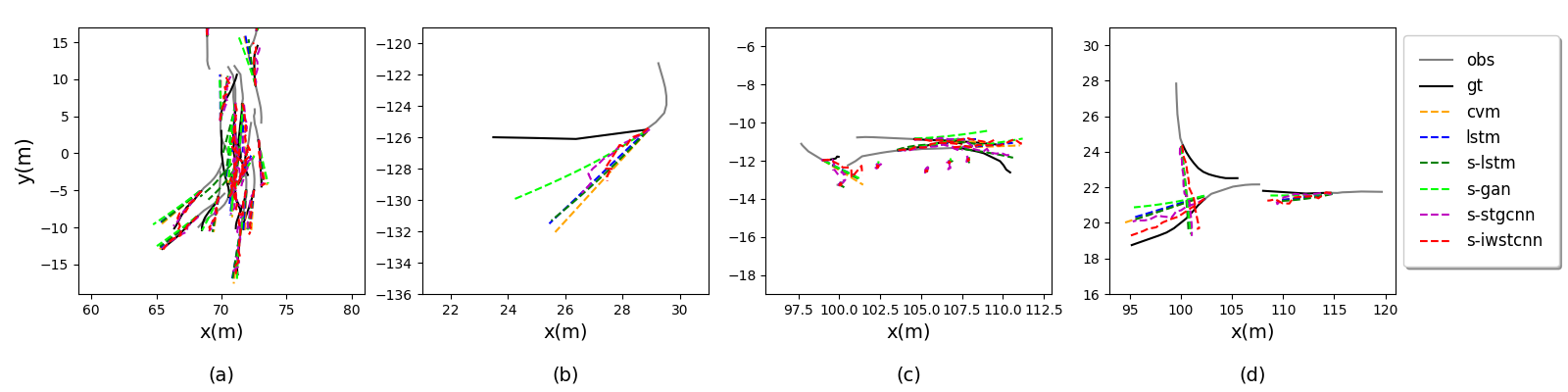} 
\end{center}
  \caption{The comparison of prediction results of different algorithms in the scenarios that are difficult to predict. (a) Trajectory prediction in a densely populated scenario. Our Social-IWSTCNN manages to capture the social interactions and outperforms other methods. (b) Failure in predicting individual changing direction suddenly, (c) failure in predicting individual changing speed suddenly, and (d) failure in collision avoidance. In (b), (c), and (d), none of the algorithms succeed in predicting the trajectories correctly, because of the lack of sufficient information.}
\label{fig:figure4}
\end{figure*}

In Fig.~\ref{fig:figure4}, we compare the results in the scenarios that are difficult to predict. The ability of dealing with dense groups is shown in Fig.~\ref{fig:figure4}(a). We find that in real-world urban scenarios, it is difficult to precisely predict the pedestrians in a dense crowd. Still, in this dense scenario, our algorithm manages to capture the interaction information between pedestrians and outperforms other methods with the learned social interaction weights. However, there are still some challenges for the Social-IWSTCNN. Fig.~\ref{fig:figure4}(b) shows that for an individual walking scenario, when the pedestrian changes his/her direction suddenly, our algorithm cannot predict the future trajectory well. This is because there is not much social interaction information for such situations. Fig.~\ref{fig:figure4}(c) shows the case that the pedestrian on the left suddenly changes his/her speed, and we fail to predict it. Fig.~\ref{fig:figure4}(d) shows that all algorithms fail when an individual needs to turn or stop to avoid a collision. In this case, if we want to predict the future trajectory more precisely, we need to include other information such as the interactions with vehicles and the scene. Images from the sensors such as cameras and LiDARs could also aid the prediction.

\section{Conclusions}
\label{sec:Conclusion}
In this paper, we presented the Social-IWSTCNN algorithm for pedestrian trajectories prediction, which outperforms major SOTA approaches while significantly reducing the computational cost of the best approach (Social-STGCNN). We proposed a novel structure, the Social Interaction Extractor, to extract the spatial and social interaction features effectively and efficiently. 
We compared the SOTA algorithms on Waymo Open Dataset to validate whether our algorithm is effective for urban traffic scenarios. While our approach has already shown decent performance in typical scenarios in the Waymo Open Dataset, it is still challenging to accurately predict the future trajectories for sparsely crowded scenarios. Therefore, further work on such scenarios is still needed. More information such as the interaction with the vehicles and the environment could also be included in the future to support the development of vehicle automation.

\addtolength{\textheight}{-3cm}   


\section*{ACKNOWLEDGMENT}
This research is funded by the European research project ``SHAPE-IT – Supporting the Interaction of Humans and Automated Vehicles: Preparing for the Environment of Tomorrow’’. This project has received funding from the European Union’s Horizon 2020 research and innovation programme under the Marie Skłodowska-Curie grant agreement 860410.



\bibliographystyle{./bibliography/IEEEtran}
\bibliography{./bibliography/IEEEexample}

\end{document}